\documentclass[conference]{IEEEtran}
\IEEEoverridecommandlockouts
\usepackage{hyperref}
\usepackage{graphicx}
\usepackage{amsfonts}           
\usepackage{cite}               
\usepackage{amsmath,amssymb,bm} 
\usepackage{mathtools}          
\usepackage{siunitx}            
\usepackage{nicefrac}           
\usepackage{booktabs}           
\usepackage{subcaption}         
\def\BibTeX{{\rm B\kern-.05em{\sc i\kern-.025em b}\kern-.08em
    T\kern-.1667em\lower.7ex\hbox{E}\kern-.125emX}}
\begin{document}

\title{Feature-Based Lie Group Transformer for Real-World Applications
}

\author{
Takayuki Komatsu$^{1}$, Yoshiyuki Ohmura$^{1}$, Kayato Nishitsunoi$^{1}$, and Yasuo Kuniyoshi$^{1,2}$
\thanks{
$^{1}$Graduate School of Information Science and Technology, The University of Tokyo, Tokyo, Japan.
$^{2}$Next Generation Artificial Intelligence Research Center (AI Center), The University of Tokyo, Tokyo, Japan.
{\tt\small \{komatsu, ohmura, nishitsunoi, kuniyoshi\}@isi.imi.i.u-tokyo.ac.jp}
This paper is partly based on results obtained under a Grant-in-Aid for Scientific Research (A) JP22H00528.
}
}

\maketitle

\begin{abstract}
The main goal of representation learning is to acquire meaningful representations from real-world sensory inputs without supervision.
Representation learning explains some aspects of human development.
Various neural network (NN) models have been proposed that acquire empirically good representations.
However, the formulation of a good representation has not been established.
We recently proposed a method for categorizing changes between a pair of sensory inputs.
A unique feature of this approach is that transformations between two sensory inputs are learned to satisfy algebraic structural constraints.
Conventional representation learning often assumes that disentangled independent feature axes is a good representation; however, we found that such a representation cannot account for conditional independence.
To overcome this problem, we proposed a new method using group decomposition in Galois algebra theory.
Although this method is promising for defining a more general representation, it assumes pixel-to-pixel translation without feature extraction, and can only process low-resolution images with no background, which prevents real-world application.
In this study, we provide a simple method to apply our group decomposition theory to a more realistic scenario by combining feature extraction and object segmentation.
We replace pixel translation with feature translation and formulate object segmentation as grouping features under the same transformation.
We validated the proposed method on a practical dataset containing both real-world object and background.
We believe that our model will lead to a better understanding of human development of object recognition in the real world.
\end{abstract}


\section{Introduction}\label{sec:introduction}

Humans can recognize essential information from sensory input, such as the shape, position, and pose of objects.
These abilities are acquired through development during infancy. 
Representation learning is an AI field that aims to learn good representations of sensory input without supervision.
Such a machine learning model can serve as a model for human cognitive development \cite{Takada2021, Nishitsunoi2024}.

An intuitive definition of good representation was proposed in the early years of research on the subject: a representation should consist of disentangled components \cite{Bengio2013}.
Mainstream representation learning methods focus on statistical independence among scalar components \cite{Higgins2017, Chen2016, Yang2023}.
They successfully learn information such as object size and uniaxial position; however, they still suffer from over-splitting a single representation into multiple components or mixing multiple representations into a single component.
Recently, Higgins et al. \cite{Higgins2018} proposed a more general definition of disentanglement based on symmetric transformations in abstract algebra.
However, they did not propose a specific learning method, and the connection between theory and method remains unresolved.

To address this problem, we proposed learning to categorize changes between sensory inputs using algebraic structural constraints.
Then we focused on algebraic independence \cite{Ohmura2025, Nishitsunoi2024}, which is a generalization of independence in mathematics \cite{Simpson2018}.
The main condition of algebraic independence is commutativity, that is, combined transformations have the same result regardless of their order: $a \circ b=b \circ a$.
Our methods successfully categorize transformations into independent vector transformations, such as color and shape transformations.
However, commutativity is not sufficient, even though it holds for independence used in almost all existing representation learning methods.
For example, translation and rotation are not commutative, but rather conditionally independent because translation affects the center of rotation.
To overcome such noncommutative transformations, it is necessary to extend from independence to conditional independence.
However, no method has been proposed to categorize conditionally independent transformations.

Recently, we proposed a transformation categorization method based on conditional independence \cite{Nishitsunoi2025}.
We focused on group decomposition in Galois theory \cite{Singh1999}, which can generalize commutativity, and combined it with Lie group transformation learning \cite{Takada2022}.
Shape-invariant geometric transformations are learned from sequential images of a moving object.
This method successfully categorized conditionally independent transformations.

Although our theory is promising for defining a more general representation, we formulate the transformations as pixel-to-pixel translations and consider only low-resolution, no-background images \cite{Takada2021,Takada2022,Takatsuki2023,Nishitsunoi2025}.
A gap remains for real-world applications because real-world images are usually high-resolution, noisy, and contain backgrounds.
For such images, it is reasonable to use feature extraction based on neural networks (NNs) \cite{Krizhevsky2012}, which achieves excellent information compression and robustness \cite{Hendrycks2019}, and is widely used in existing representation learning methods \cite{Higgins2017, Chen2016, Yang2023}.

\begin{figure*}[t]\centering\includegraphics[width=\linewidth]{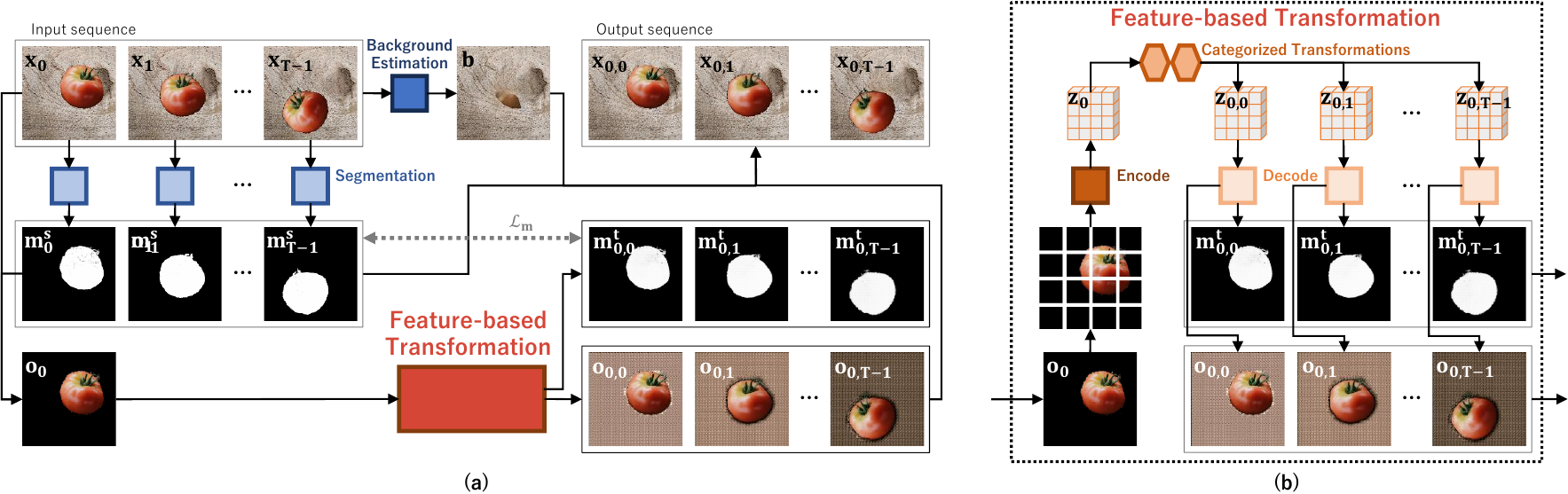}\caption{
\textbf{(a)} Overview of our proposed method.
An input sequence consists of $T$ scene images $\{\mathbf{x}_i\}_{i=0}^{T-1}$.
A background image $\mathbf{b}$ and object masks $\{\mathbf{m}^s_i\}_{i=0}^{T-1}$ are inferred from the scene images $\{\mathbf{x}_i\}_{i=0}^{T-1}$.
A head object image $\mathbf{o}_0$ is segmented from the head scene image $\mathbf{x}_0$ using the object mask $\mathbf{m}^s_0$.
Transformed object images $\{\mathbf{o}_{0,i}\}_{i=0}^{T-1}$ and transformed masks $\{\mathbf{m}^t_{0,i}\}_{i=0}^{T-1}$ are obtained from the head object image $\mathbf{o}_0$.
A mask reconstruction loss $\mathcal{L}_m$ constrains the object mask $\mathbf{m}^s_i$ to match the transformed mask $\mathbf{m}^t_{0,i}$.
Finally, an output sequence $\{\mathbf{x}_{0,i}\}_{i=0}^{T-1}$ is obtained from the object masks $\{\mathbf{m}^s_i\}_{i=0}^{T-1}$, transformed object images $\{\mathbf{o}_{0,i}\}_{i=0}^{T-1}$, and background image $\mathbf{b}$.
\textbf{(b)} Overview of feature-based transformation.
The head object $\mathbf{o}_0$ is divided into $N \times N$ patches.
Each patch is encoded into a feature vector, which yields a head feature image $\mathbf{z}_0$ that consists of $N \times N$ feature vectors.
Transformed feature images $\{\mathbf{z}_{0,i}\}_{i=0}^{T-1}$ are obtained from the head feature image $\mathbf{z}_0$ using a combination of two categorized transformations.
Finally, the transformed object images $\{\mathbf{o}_{0,i}\}_{i=0}^{T-1}$ and transformed masks $\{\mathbf{m}^t_{0,i}\}_{i=0}^{T-1}$ are obtained from the transformed feature images $\{\mathbf{z}_{0,i}\}_{i=0}^{T-1}$.
}\label{fig:1}\end{figure*}

In this study, we combine feature extraction and object segmentation with newly developed group decomposition.
We replace pixel translation with feature translation.
We then formulate object segmentation as grouping features under the same transformation into a single object.
The overview of the proposed method is shown in Fig.~\ref{fig:1}.

We validated the proposed method on high-resolution images containing real-world object and background.
For the first time, we achieved the simultaneous unsupervised categorization of feature-based Lie group transformation, and object segmentation on a real-world object and background.
Our model contributes to a model of how humans develop the ability to detect and recognize objects in the real world.

\section{Transformation Categorization}

In this section, we briefly review the formulation of transformation categorization provided by our group \cite{Nishitsunoi2025}.

\subsection{Normal subgroups}

Given a group $G$, the normal subgroup $N \subset G$ is defined as follows:
\begin{equation}
    gN=Ng, \quad \forall g \in G,
\end{equation}
where $gN$ and $Ng$ are the left and right cosets of $N$ in $G$,
\begin{align}
    gN &= \{g \circ n|n\in N\}, \quad \forall g \in G, \\
    Ng &= \{n \circ g|n\in N\}, \quad \forall g \in G.
\end{align}
Commutativity requires $g \circ n=n \circ g$; however, the normal subgroup allows elements $n$ to be different on the right and left-hand sides.
Therefore, normal subgroups are more general than commutativity.
The normal subgroup $N$ is known as a well-defined partition of $G$, which Galois referred to as a proper decomposition \cite{Singh1999}.

Normal subgroups can be obtained from group homomorphisms.
A homomorphism $f: G \to G'$ is defined as a function between groups $G$ and $G'$ that satisfies
\begin{equation}
    f(g_1 \circ g_2) = f(g_1) \cdot f(g_2), \quad \forall g_1, g_2 \in G.
    \label{eq:group_homomorphism}
\end{equation}
Then the kernel of the homomorphism $f$ is defined as follows:
\begin{equation}
    \ker (f) := \{g \in G|f(g) = e\},
    \label{eq:group_kernel}
\end{equation}
where $e$ is the identity element of $G'$.
The kernel of the homomorphism $f$ is the normal subgroup of $G$.
Thus, we formulate transformations as group elements and then categorize transformations into $\ker (f)$ and the other.

\subsection{Formulation of transformation categorization}

We consider transformations in sequences where each sequence consists of $T$ images $\{\mathbf{o}_i\}_{i=0}^{T-1}$.
Each image $\mathbf{o}_i$ contains a single object and no background.
The object pixels are non-zero values and we assume that the background pixels are zero.
We let $\mathbf{o}_i$ denote an object image.
All objects in the sequence have the same shape and color, and differ only in position and orientation.
An index $i$ reflects the elapsed timestep and the sequence $\mathbf{o}_{0:T-1}$ expresses the object that moves as timestep $i$ increases.
Object motion is a combination of two geometric transformations, such as object-centered rotation and translation.

We let $i$-$j$ transformation denote the transformation from the $i$-th object image $\mathbf{o}_i$ to the $j$-th object image $\mathbf{o}_j$.
We aim to learn two transformations $g$ and $v$, and to represent the $i$-$j$ transformation by a combination of these two transformations.
We let $g_{i,j}$ and $v_{i,j}$ denote the transformations used in the $i$-$j$ transformation.
We define a transformed object image $\mathbf{o}_{i,j}$ as follows:
\begin{equation}
    \mathbf{o}_{i,j} = (g_{i,j} \circ v_{i,j})\mathbf{o}_i.
    \label{eq:object_motion}
\end{equation}
Two transformations $g_{i,j}$ and $v_{i,j}$ are required to match the transformed object image $\mathbf{o}_{i,j}$ with the ground truth object image $\mathbf{o}_j$.

These two transformations $g$ and $v$ are also required to be categorized appropriately based on group decomposition theory.
We assume a group $G$ whose elements are the combinations of two transformations $(g \circ v)$.
We consider a homomorphism $f$ that satisfies
\begin{equation}
    f(g \circ v) = v.
    \label{eq:define_f}
\end{equation}
The kernel of this homomorphism $f$ coincides with the set of $g$ as follows:
\begin{align}
    \ker (f) &= \{g \circ v \in G| f(g \circ v) = e\} \notag \\
    &= \{g \circ v \in G|v = e\} \notag \\
    &= \{g \in G\}.
\end{align}
Then we obtain the set of $g$ as the normal subgroup $N$ and $v$ as a transformation that cannot be expressed by $g$ alone.

When the homomorphism $f$ satisfies both Eqs.~\ref{eq:group_homomorphism} and \ref{eq:define_f}, we obtain the equation for $v$ as follows, with $(g_c \circ v_c) = (g_a \circ v_a) \circ (g_b \circ v_b)$,
\begin{align}
    f((g_a \circ v_a) \circ (g_b \circ v_b)) &= f(g_a \circ v_a) \cdot f(g_b \circ v_b), \notag \\
    f(g_c \circ v_c) &= f(g_a \circ v_a) \cdot f(g_b \circ v_b), \notag \\
    v_c &= v_a \cdot v_b.
    \label{eq:v_homomorphism}
\end{align}
Therefore, we aim to constrain $v$ to satisfy Eq.~\ref{eq:v_homomorphism}.

\subsection{Lie group transformation}\label{sec:lie_group_transformation}

In our previous study \cite{Nishitsunoi2025}, we focus on geometric transformations such as translations and rotations.
Therefore, we make two assumptions about transformations $g$ and $v$:
(i) they can be formulated using pixel-to-pixel translations
and (ii) they have the Lie group property because geometric transformations are known to have this property.

For a single transformation of $g$ or $v$, let $\mathbf{o}^0$ be an input image and $\mathbf{o}^t$ be an output image.
We formulate the transformation as translating a pixel value (e.g., RGB value) at coordinates $(x^0, y^0)$ of the input image $\mathbf{o}^0$ to coordinates $(x^t, y^t)$ of the output image $\mathbf{o}^t$.
We implement the transformation using NeuralODE \cite{Chen2018}, which performs the integral and is guaranteed to satisfy the Lie group property.
We obtain a pixel shift ($\Delta x^t$, $\Delta y^t$)=($x^t$-$x^0$, $y^t$-$y^0$) as follows:
\begin{equation}
    \begin{bmatrix}
        \Delta x^t \\
        \Delta y^t
    \end{bmatrix}
    = \int_{0}^{\lambda} f\left(\begin{bmatrix}
        x(t) \\
        y(t)
    \end{bmatrix}\right)
    dt,
\end{equation}
\begin{equation}
    f \left(\begin{bmatrix}
        x(t) \\
        y(t)
    \end{bmatrix}\right)
    = A\begin{bmatrix}
        x(t) \\
        y(t)
    \end{bmatrix}
    + \mathbf{b} + \mathbf{c},
\end{equation}
where $\lambda \in \mathbb{R}$ is the upper or lower limit of the integral, $A \in \mathbb{R}^{2 \times 2}$ is a matrix, $\mathbf{b} \in \mathbb{R}^{2 \times 1}$ is a bias, and $\mathbf{c} \in \mathbb{R}^{2 \times 1}$ is a conditional variable.
$A$ and $\mathbf{b}$ are the same for all sequences and all $i$-$j$ transformations.
By contrast, $\lambda$ and $\mathbf{c}$ are assigned for each sequence and each $i$-$j$ transformation.
Then, $\lambda$ reflects the amount of transformation and $\mathbf{c}$ reflects a factor other than $\lambda$, such as the rotation center and translation direction.
All $A, \mathbf{b}, \lambda$, and $\mathbf{c}$ are provided for both $g$ and $v$.
Particulary for the transformation parameters $\lambda$ and $\mathbf{c}$, we denote the parameters used in $i$-$j$ transformation as $\lambda^g_{i,j}$, $\mathbf{c}^g_{i,j}$, $\lambda^v_{i,j}$, and $\mathbf{c}^v_{i,j}$.
Transformations $g_{i,j}$ and $v_{i,j}$ can be written as $g(\lambda^g_{i,j}, \mathbf{c}^g_{i,j})$, and $v(\lambda^v_{i,j}, \mathbf{c}^v_{i,j})$.

We let a pixel value at coordinates $(x, y)$ of an image $\mathbf{o}$ denote $\mathbf{o}[x, y]$.
We obtain the pixel value of the output image $\mathbf{o}^t[x^t, y^t]$ using pixel values in the neighborhood of the corresponding coordinates $(x^0, y^0)$ of the input image $\mathbf{o}^0$ as follows:
\begin{equation}
    \mathbf{o}^t[x^t, y^t] = \sum_{x,y}^{W,H} w \mathbf{o}^0[x, y],\label{eq:weighted_interpolation}
\end{equation}
where $w \in \mathbb{R}$ is a weight to reflect only the value of neighboring pixels defined as follows, with $d_x = |x - x^0|$ and $d_y = |y - y^0|$:
\begin{equation}
    w = \begin{cases}
        (1 - d_x)(1 - d_y) & \text{if } d_x < 1 \text{ and } d_y < 1 \\
        0 & \text{otherwise.}\label{eq:interpolation_weight}
    \end{cases}
\end{equation}
This means that transformations $g$ and $v$ do not change the pixel values, except for interpolation using the neighboring pixels.

\subsection{Training method}\label{sec:learning_model}

In the previous study, for object motion, we assume that the amount of transformation per timestep and the translation direction are constant for each sequence.
Then we mainly focus on learning the $0$-$i$ transformations that are formulated using $0$-$1$ transformation parameters as follows:
\begin{align}
    g_{0,i} = g(i \times \lambda^g_{0,1}, \mathbf{c}^g_{0,1}), \\
    v_{0,i} = v(i \times \lambda^v_{0,1}, \mathbf{c}^v_{0,1}).
    \label{eq:stationary_transformation}
\end{align}

We use several loss functions.
We use the first loss $\mathcal{L}_r$ to match the transformed image $\mathbf{o}_{0,i}$ and ground truth image $\mathbf{o}_i$.
The reconstruction loss $\mathcal{L}_r$ is defined using the mean squared errors (MSEs) as follows:
\begin{equation}
    \mathcal{L}_{r} = \sum_{i=0}^{T-1} MSE(\mathbf{o}_i, \mathbf{o}_{0, i}).
    \label{eq:pixel_base_recons_loss}
\end{equation}

We use the second loss $\mathcal{L}_h$ to acquire the homomorphism $f$, which is derived from Eq.~\ref{eq:v_homomorphism}.
The homomorphism loss $\mathcal{L}_h$ is defined as follows, by setting $v_a = v_{0,1}$, $v_b = v_{1,2}$, and $v_c = v_{0,2}$:
\begin{equation}
    \mathcal{L}_h = MSE(v_{0,2}\mathbf{o}_0, (v_{0,1} \cdot v_{1,2})\mathbf{o}_0).
\end{equation}
Regarding the transformation $v_{1,2}$, the transformed object image $\mathbf{o}_{1,2}$ is required to match the ground truth object image $\mathbf{o}_2$.
A third loss $\mathcal{L}_{r2}$ is defined as follows:
\begin{equation}
    \mathcal{L}_{r2} = MSE(\mathbf{o}_2, \mathbf{o}_{1,2}).
\end{equation}

We use the fourth loss $\mathcal{L}_{s}$ to ensure that the homomorphism $f$ satisfies Eq.~\ref{eq:define_f}, where the homomorphism $f$ identifies $g$ from the combined transformation $(g \circ v)$.
We perform self-supervised learning (SSL) where either $g$ or $v$ is replaced by an identity transformation $e$.
We generate two sequences based on Eq.~\ref{eq:stationary_transformation}, replacing either $g$ or $v$ with the identity transformation $e$, as follows:
\begin{align}
    \mathbf{o}^{g \to e}_{0:T-1} = \{ v(i \times \lambda^v_{0,1}, \mathbf{c}^v_{0,1})\mathbf{o}_0 \}_{i=0}^{T-1}, \\
    \mathbf{o}^{v \to e}_{0:T-1} = \{ g(i \times \lambda^g_{0,1}, \mathbf{c}^g_{0,1})\mathbf{o}_0 \}_{i=0}^{T-1}.
\end{align}
We prepare an estimator $I$ of the transformation parameters as follows:
\begin{equation}
    I(\mathbf{o}_{0:T-1}) = (\lambda^g_{0,1}, \mathbf{c}^g_{0,1}, \lambda^v_{0,1}, \mathbf{c}^v_{0,1}).
\end{equation}
Then the SSL loss $\mathcal{L}_s$ is defined as the parameter estimation error from the two sequences as follows:
\begin{align*}
    \mathcal{L}_s = 
    &MSE(I(\mathbf{o}^{g \to e}_{0:T-1}), (0, \mathbf{0}, \lambda^v_{0,1}, \mathbf{c}^v_{0,1})) \\
    + &MSE(I(\mathbf{o}^{v \to e}_{0:T-1}), (\lambda^g_{0,1}, \mathbf{c}^g_{0,1}, 0, \mathbf{0})).
\end{align*}
We implement the parameter estimator $I$ using a convolutional NN (CNN) and long short-term memory (LSTM) \cite{Hochreiter1997}.

We assume that transformations $g$ and $v$ are isometric and use the fifth loss $\mathcal{L}_i$ to satisfy this assumption.
We pair the pixels in the image and compute the relative distances of the coordinates to obtain a distance matrix $M \in \mathbb{R}^{HW \times HW}$, where $H$ and $W$ are the height and width of the image.
Using the matrix before transformation $M_0$, and after a transformation $M^g_{0, i}$ and $M^v_{0, i}$, we define the isometry loss $\mathcal{L}_i$ as follows:
\begin{equation}
    \mathcal{L}_i = \sum_{i=0}^{T-1} \{MSE(M_0, M^g_{0, i}) + MSE(M_0, M^v_{0, i})\}.
\end{equation}

The sixth loss $\mathcal{L}_c$ regularizes the conditional variables $\mathbf{c}$ so that various combinations of $\lambda$ and $\mathbf{c}$ cannot represent the same transformation:
\begin{equation}
    \mathcal{L}_c = \|\mathbf{c}^g_{0,1}\|_2 + \|\mathbf{c}^g_{1,2}\|_2 + \|\mathbf{c}^v_{0,1}\|_2 + \|\mathbf{c}^v_{1,2}\|_2.
\end{equation}

To summarize, the total loss function is as follows:
\begin{equation}
    \mathcal{L} = \alpha \mathcal{L}_r + \beta \mathcal{L}_{r2} + \gamma \mathcal{L}_h + \delta \mathcal{L}_s + \epsilon \mathcal{L}_i + \zeta \mathcal{L}_c,
    \label{eq:loss_all}
\end{equation}
where $\alpha, \beta, \gamma, \delta, \epsilon, \zeta \in \mathbb{R}$ are weights for each loss.

\begin{figure}[t]\centering\includegraphics[width=0.97\linewidth]{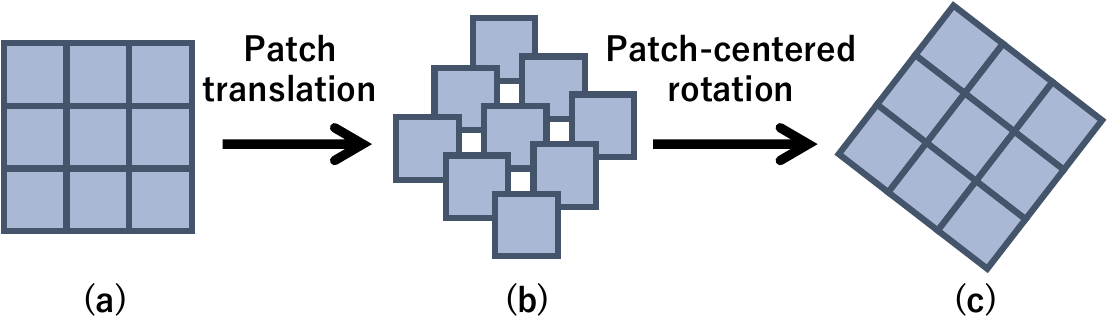}\caption{
Schematic of object transformation.
\textbf{(a)} An object before transformation.
\textbf{(b)} A transformed object with patch translation.
\textbf{(c)} A transformed object with patch translation and patch-centered rotation.
}\label{fig:5}\end{figure}

The previous method has two problems.
First, the transformations are pixel-to-pixel translations and the method does not consider feature extraction.
It is difficult to process real-world images, which are high-resolution and noisy.
Second, transformations are performed on all pixels in the image.
When the image contains both the background and object, pixel translation must be performed only on the pixels of the moving object.
However, the method does not consider the classification of object pixels and background pixels.
Because of these two problems, we use only simple images with low resolution and no background.
In this study, we combine feature extraction and object segmentation with the previous method, and retain the formulation based on group decomposition theory described in this section.

\section{Method}\label{sec:method}
The transformation consists of two processes:
(i) feature-based transformation of the object image
and (ii) separation of the object from the background.

\subsection{Feature-based transformation}\label{sec:feature_based_transformation}

We consider the object image $\mathbf{o}_i$ whose background pixels are zero.
We divide the object image $\mathbf{o}_i$ into $N \times N$ patches, where the size of each patch is $(H/N) \times (W/N)$.
We denote a single patch as $\mathbf{p}_i[x,y]$.
We prepare an NN model that infers a feature vector $\mathbf{z}_i[x,y] \in \mathbb{R}^Z$ from the single patch $\mathbf{p}_i[x,y]$.
We process each patch using the NN model to obtain a feature image $\mathbf{z}_i \in \mathbb{R}^{N \times N \times Z}$.
We define an encoder $E$, which consists of the image division and the NN model, as follows:
\begin{equation}
    \mathbf{z}_i = E(\mathbf{o}_i).
\end{equation}

We simply replace the translations on $H \times W$ pixels in Section \ref{sec:lie_group_transformation} with translations on $N \times N$ feature vectors.
We perform feature translations using interpolation as follows, similar to Eq.~\ref{eq:weighted_interpolation}:
\begin{equation}
    \mathbf{z}^t[x^t, y^t] = \sum_{x,y}^{N,N} w \mathbf{z}^0[x, y],
\end{equation}
where $\mathbf{z}^0$ is an input feature image and $\mathbf{z}^t$ is an output feature image, and we obtain a weight $w$ similar to Eq.~\ref{eq:interpolation_weight}.
We obtain a transformed feature image $\mathbf{z}_{i,j}$ as follows:
\begin{equation}
    \mathbf{z}_{i,j} = (g_{i,j} \circ v_{i,j})\mathbf{z}_i.
\end{equation}
Although the feature translations can represent patch translations, feature interpolations are not sufficient to represent patch-centered rotations, which are necessary to perform object rotation as shown in Fig.~\ref{fig:5}.
This problem does not occur in pixel translation because pixel values are rotation-invariant.
For performing patch-centered rotation, it is also necessary to transform feature vectors.

To achieve this, instead of preparing a transformer for a translated feature vector $\mathbf{z}^t[x^t, y^t]$, we propose a method to rotate the translated feature vector $\mathbf{z}_{i,j}[x, y]$ in a decoder by simply providing the decoder with the transformation parameters $(\mathbf{c}^g_{i,j}, \mathbf{c}^v_{i,j}, \lambda^g_{i,j}, \lambda^v_{i,j})$.
We define a conditioned decoder $D_{i,j}$ as follows:
\begin{equation}
    \mathbf{o}_{i,j} = D_{i,j}(\mathbf{z}_{i,j}) = D(\mathbf{z}_{i,j}, \mathbf{c}^g_{i,j}, \mathbf{c}^v_{i,j}, \lambda^g_{i,j}, \lambda^v_{i,j}).
\end{equation}
Using information about the transformation, the conditioned decoder $D_{i,j}$ can output the transformed object image $\mathbf{o}_{i,j}$ that reflects the transformation of patches.
We broadcast the transformation parameters $(\mathbf{c}^g_{i,j}$, $\mathbf{c}^v_{i,j}$, $\lambda^g_{i,j}$, $\lambda^v_{i,j})$, 
concatenate them into the transformed feature image $\mathbf{z}_{i,j}$, 
and acquire the transformed object image $\mathbf{o}_{i,j}$ using a CNN.

Finally, the $i$-$j$ transformation is formulated as follows:
\begin{equation}
    \mathbf{o}_{i,j} = (E \circ g_{i,j} \circ v_{i,j} \circ D_{i,j})\mathbf{o}_i.
    \label{eq:feature_transformation_output}
\end{equation}

We continue to use the same loss functions in Eq.~\ref{eq:loss_all}.
We compute $\mathcal{L}_r$, $\mathcal{L}_{r2}$, and $\mathcal{L}_h$ by replacing the $i$-$j$ transformation from Eq.~\ref{eq:object_motion} to Eq.~\ref{eq:feature_transformation_output}.

Our proposed method retains the formulation of transformations $g$ and $v$ based on group decomposition theory.
We successfully combine our previous transformation categorization method with feature extraction.
Transformations $g$ and $v$ do not translate $H \times W$ raw pixels, but rather $N \times N$ feature vectors.
Therefore, we can reduce computational cost for translations on high-resolution images where $H$ and $W$ are much larger than $N$.

\subsection{Feature-transformation-based object segmentation}\label{sec:feature_transformation_based_object_segmentation}

We consider a sequence $\mathbf{x}_{0:T-1} = \{\mathbf{x}_i\}_{i=0}^{T-1}$, where $\mathbf{x}_i \in \mathbb{R}^{H \times W \times 3}$ is a scene image containing both the background and object.
We assume that only the object moves and the background does not change in the sequence.
We aim to separate image regions under the same transformation from the stationary background and group them as a single object.
We group features under the same feature transformation $(g_{i,j} \circ v_{i,j})$ as an object.

We formulate object segmentation as a multiplication of an object mask $\mathbf{m}^s_i \in \mathbb{R}^{H \times W \times 1}$ and the scene image $\mathbf{x}_i$ as follows:
\begin{equation}
    \mathbf{o}_i = \mathbf{m}^s_i \mathbf{x}_i.
\end{equation}
The object mask $\mathbf{m}^s_i$ takes values between $0$ and $1$, where $1$ indicates an object pixel and $0$ indicates a background pixel.
By multiplying the background pixels by $0$, we obtain the object image $\mathbf{o}_i$ with zero background pixels, which we assume in the previous section.
Then we can apply transformations in Eq.~\ref{eq:feature_transformation_output} to the object image $\mathbf{o}_i$ and obtain the transformed object image $\mathbf{o}_{i, j}$.
We call $(1-\mathbf{m}^s_i)$ a background mask.
We obtain the object mask $\mathbf{m}^s_i$ from the scene image $\mathbf{x}_i$ using a segmenter $S$ as follows:
\begin{equation}
    \mathbf{m}^s_i = S(\mathbf{x}_i).
    \label{eq:object_segmentation_mask}
\end{equation}
We implement the segmenter $S$ using U-Net \cite{Ronneberger2015} and instance normalization \cite{Ulyanov2016} as well as MONet \cite{Burgess2019}, which is an unsupervised segmentation method.
We also normalize all the $HW$ pixels of the object mask $\mathbf{m}^s_i$ so that the maximum value is $1$ and minimum value is $0$: $\mathbf{m}^s_i \leftarrow (\mathbf{m}^s_i - \min(\mathbf{m}^s_i))/(\max(\mathbf{m}^s_i) - \min(\mathbf{m}^s_i))$.

The object mask $\mathbf{m}^s_i$ should represent image region under the transformation $(g_{i,j} \circ v_{i,j})$.
However, as shown in Eq.~\ref{eq:object_segmentation_mask}, the transformation $(g_{i,j} \circ v_{i,j})$ is not involved in the inference of the object mask $\mathbf{m}^s_i$.
It is necessary to constrain the object masks of different timesteps $\mathbf{m}^s_i$ and $\mathbf{m}^s_j$ to be under the same transformation $(g_{i,j} \circ v_{i,j})$.
To achieve this, we introduce a new mask generated from the transformed feature image $\mathbf{z}_{i, j}$ and constrain the object mask $\mathbf{m}^s_i$ to match this new mask.
We let this mask denote a transformed mask $\mathbf{m}^t_{i, j}$ and obtain this mask using the decoder $D_{i, j}$ as follows:
\begin{equation}
    (\mathbf{o}_{i, j}, \mathbf{m}^t_{i, j}) = D_{i, j}(\mathbf{z}_{i, j}).
\end{equation}
We implement the decoder $D_{i, j}$ to output a $4$-channel image instead of a $3$-channel image.
Then we split the $4$-channel image into the transformed object image $\mathbf{o}_{i, j}$ and transformed mask $\mathbf{m}^t_{i, j}$.
We normalize the transformed mask $\mathbf{m}^t_{i, j}$, in addition to the object mask $\mathbf{m}^s_i$.
We define the mask reconstruction loss $\mathcal{L}_m$ as follows:
\begin{equation}
    \mathcal{L}_m = \sum_{i=0}^{T-1} MSE(\mathbf{m}^s_i, \mathbf{m}^t_{0, i}).
\end{equation}

In addition to the transformed object image $\mathbf{o}_{i, j}$, we also consider a background image $\mathbf{b}$.
We obtain the background image $\mathbf{b}$ from the sequence $\mathbf{x}_{0:T-1}$ using a background estimator $B$ as follows:
\begin{equation}
    \mathbf{b} = B(\mathbf{x}_{0:T-1}).
\end{equation}
We concatenate the sequences in the image channel direction and output the background $\mathbf{b}$ using a CNN.

We define a transformed scene image $\mathbf{x}_{i,j}$ as follows:
\begin{equation}
    \mathbf{x}_{i,j} = \mathbf{m}^s_i \mathbf{o}_{i,j} + (1 - \mathbf{m}^s_i) \mathbf{b}.
\end{equation}
When the transformed object image $\mathbf{o}_{i,j}$, background image $\mathbf{b}$, and object mask $\mathbf{m}^s_i$ are appropriate, the transformed scene image $\mathbf{x}_{i,j}$ matches the ground truth scene image $\mathbf{x}_i$.

We modify the reconstruction losses $\mathcal{L}_r$ and $\mathcal{L}_{r2}$  to account for a object and background.
An approach may exist to compute the MSE between the transformed scene image $\mathbf{x}_{i,j}$ and the ground truth scene image $\mathbf{x}_i$: $MSE(\mathbf{x}_i, \mathbf{x}_{i,j})$.
However, this MSE does not consider the distance between the scene and object image $MSE(\mathbf{x}_i, \mathbf{o}_{i,j})$, and the distance between the scene and background image $MSE(\mathbf{x}_i, \mathbf{b})$ directly.
Therefore, it may lead to inappropriate result such as segmenting the scene image $\mathbf{x}_i$ in the RGB channel direction using the object mask $\mathbf{m}^s_i$ \cite{Komatsu2024}.
To prevent this, we compute two squared errors $(\mathbf{x}_i - \mathbf{o}_{i,j})^2$ and $(\mathbf{x}_i - \mathbf{b})^2$ directly, and design mask-weighted reconstruction losses $\mathcal{L}'_r$ and $\mathcal{L}'_{r2}$ as follows:
\begin{align}
    \mathcal{L}'_r &= \sum_{i=0}^{T-1} \left\{ \mathbf{m}^s_i(\mathbf{x}_{i} - \mathbf{o}_{0,i})^2 + (1 - \mathbf{m}^s_i)(\mathbf{x}_{i} - \mathbf{b})^2 \right\}, \\
    \mathcal{L}'_{r2} &= \mathbf{m}^s_2(\mathbf{x}_{2} - \mathbf{o}_{1,2})^2 + (1 - \mathbf{m}^s_2)(\mathbf{x}_{2} - \mathbf{b})^2.
\end{align}
To minimize these losses, for each squared error and its corresponding mask, the mask values must be zero for region with large squared errors and non-zero only for region with zero squared errors.
Therefore, the object mask $\mathbf{m}^s_i$ reflects the image region of the scene image $\mathbf{x}_i$ where the transformed object image $\mathbf{o}_{0,i}$ can infer correctly, and the background mask $(1 - \mathbf{m}^s_i)$ reflects the region where the background image $\mathbf{b}$ can infer correctly.

To summarize, the total loss function is as follows:
\begin{equation}
    \mathcal{L} = \alpha \mathcal{L}'_r + \beta \mathcal{L}'_{r2} + \gamma \mathcal{L}_h + \delta \mathcal{L}_s + \epsilon \mathcal{L}_i + \zeta \mathcal{L}_c + \eta \mathcal{L}_m,
    \label{eq:loss_all_with_background}
\end{equation}
where $\alpha, \beta, \gamma, \delta, \epsilon, \zeta, \eta \in \mathbb{R}$ are the weights for each loss.

Finally, our proposed method can simultaneously learn object segmentation and transformation categorization without supervision.
Object segmentation is formulated as the grouping of features under the same transformation.

\section{Experiments}\label{sec:experiments}

We performed two experiments using different datasets of high-resolution images: 
(i) containing a synthetic object and no background (\textbf{Syn-obj}) to validate the proposed method about feature vector transformation, 
and (ii) containing a real-world object and background (\textbf{Real-obj-bg}) to validate the proposed method about object segmentation.

\begin{figure}[t]\centering
\begin{subfigure}[t]{0.459\columnwidth}\centering\fbox{\includegraphics[width=0.95\textwidth]{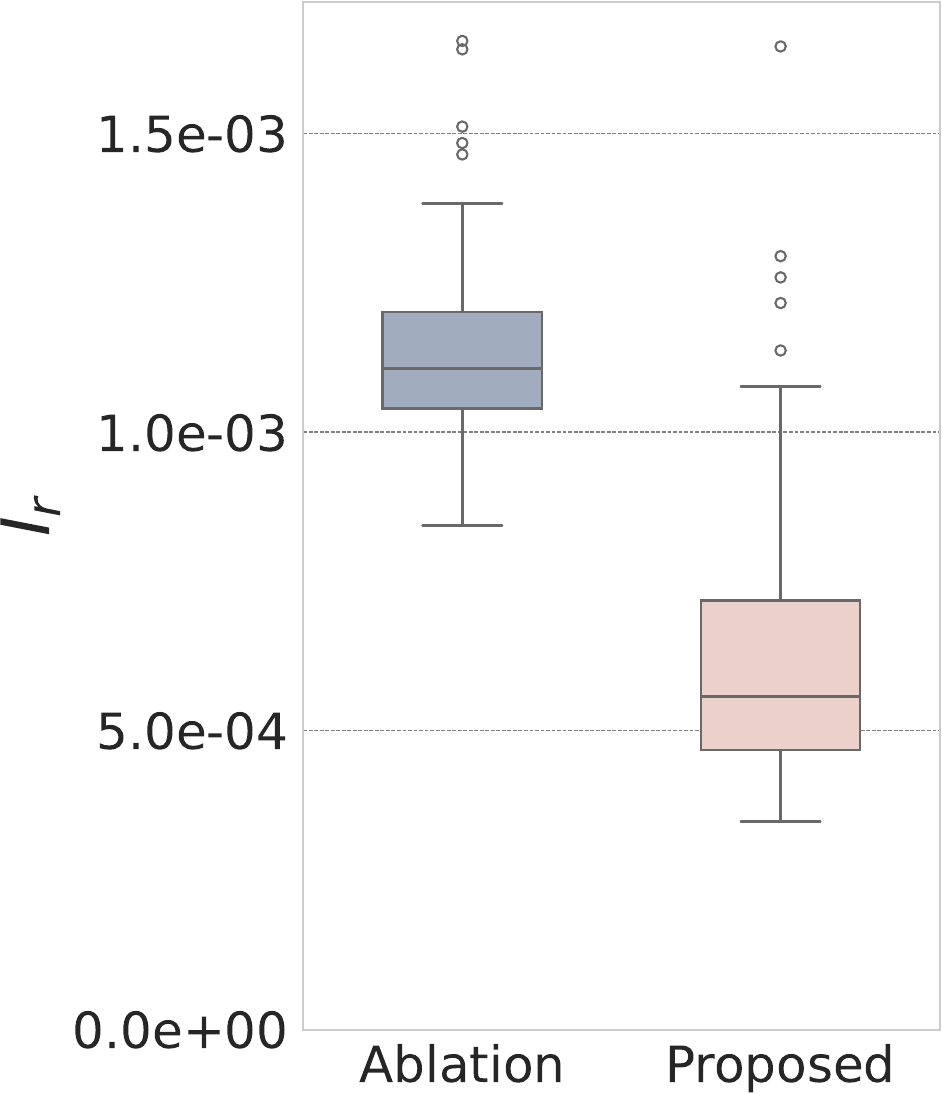}}\caption{}\label{fig:2_a}\end{subfigure}
\begin{subfigure}[t]{0.521\columnwidth}\centering\fbox{\includegraphics[width=0.95\textwidth]{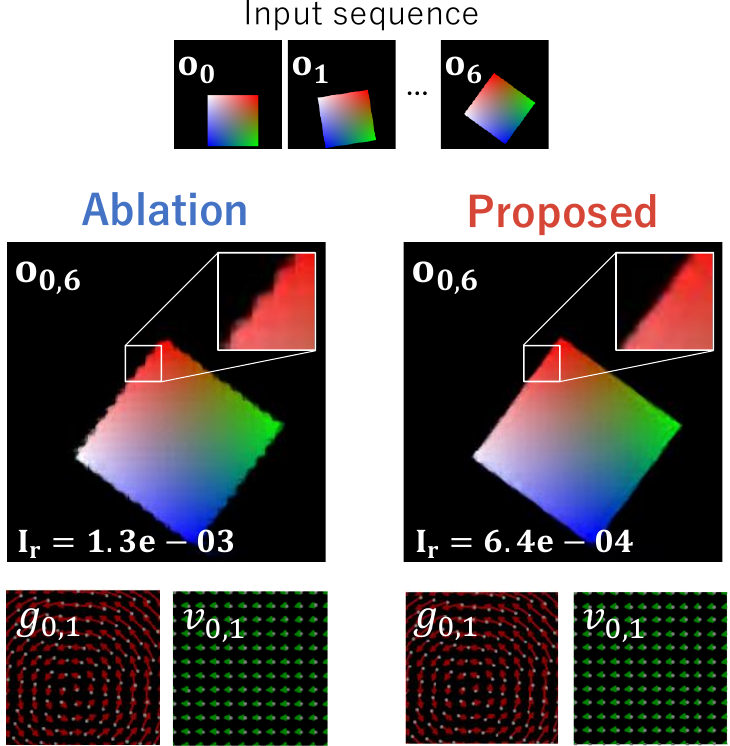}}\caption{}\label{fig:2_b}\end{subfigure}\caption{
Results of comparison experiment on \textbf{Syn-obj}. 
\textbf{(a)} Boxplot comparing the image reconstruction metric $I_r$ between the decoder without conditioning $D$ (ablation) and the conditioned decoder $D_{i,j}$ (proposed).
\textbf{(b)} Examples of transformed images $\mathbf{o}_{0,6}$ and transformations $g_{0,1}$ and $v_{0,1}$.
The image reconstruction metric $I_r$ is the metric in one sequence, not the average over all sequences.
}\label{fig:2}\end{figure}

\subsection{NN models and training settings}

In this section, we describe the NN models and their training settings, which are mostly the same in the two experiments.
To describe the NN configuration, we define $L(i, o)$ as a biased linear matrix of input dimension $i$ and output dimension $o$: 
$C_{k, s, p}(i, o)$ as a convolution of input channel $i$, output channel $o$, kernel size $k \times k$, stride $s \times s$, and padding $p \times p$: 
$TC_{k, s, p}(i, o)$ as a transposed convolution of similar parameters to those for the convolution, 
and $R$ as a ReLU activation function.

The number of patches $N$ was $32$.
The dimension of feature $Z$ was $16$.
The encoder was $L(192, 16)$.
The decoder $D_{i,j}$ was a successive process of [$L(d_{in}, 128)$, $R$, $TC_{8, 8, 0}(128, 128)$, $R$, $C_{3, 1, 1}(128, 64)$, $R$, $C_{1, 1, 0}(64, d_{out})$].
For the decoder without conditioning $D$ for ablation study, $d_{in}$ was $16$; otherwise, it was $22$.
In experiment (i), $d_{out}$ was $3$, and in experiment (ii), $d_{out}$ was $4$.
We only used the background estimator $B$ and segmenter $S$ in experiment (ii).
The configuration of $S$ was the same as that for MONet \cite{Burgess2019}.
The background estimator $B$ was a successive process of [$C_{3, 1, 1}(21, 64)$, $R$, $C_{3, 1, 1}(64, 64)$, $R$, $C_{1, 1, 0}(64, 3)$].

The CNN module of the parameter estimator $I$ was a successive process of [$C_{3, 2, 1}(3, 16)$, $R$, $C_{3, 2, 1}(16, 32)$, $R$, $C_{3, 2, 1}(32, 64)$, $R$, $C_{3, 2, 1}(64, 128)$, $R$, $C_{3, 2, 1}(128, 128)$, $R$, $C_{1, 1, 0}(128, 128)$].
The configuration of the LSTM module and the settings for transformers $g$ and $v$ were the same as in our previous method \cite{Nishitsunoi2025}.

The batch size was $5$.
The number of optimization steps was $5000$ for experiment (i) and $10000$ for experiment (ii).
For experiment (i), we set $(\alpha, \beta, \gamma, \delta, \epsilon, \zeta)$ in Eq.~\ref{eq:loss_all} to $(1, 1, 1, 0.1, 1, 0.05)$.
For experiment (ii), we set $(\alpha, \beta, \gamma, \delta, \epsilon, \zeta, \eta)$ in Eq.~\ref{eq:loss_all_with_background} to $(1, 1, 0.1, 0.01, 1, 0.005, 0.1)$.

For both experiments, we used $100$ random seeds ranging from $1$ to $100$.
The difference in random seeds affected tasks such as batch sampling and the initialization of the NNs.
We prepared a common dataset that we used across different random seeds.

The other settings were the same as those for our previous method \cite{Nishitsunoi2025}.

\subsection{Feature transformation}\label{sec:feature_transformation}

We demonstrated the effectiveness of the conditioned decoder $D_{0,i}$ on \textbf{Syn-obj}.
We compared performance with an ablation condition: a decoder $D$ whose input was only the transformed feature image $\mathbf{z}_{0,i}$.

The object in \textbf{Syn-obj} was a square colored using a gradient so that each vertex was white, red, blue, and green.
The size of the image was $256 \times 256$.
The size of the object was $120 \times 120$.
The number of images per sequence $T$ was $7$.
The number of sequences was $10$.

The transformation of the object consisted of translation and object-centered rotation.
The amount of transformation was constant over all timesteps $i$.
The initial position and amount of transformation differed among sequences, and they were determined randomly under the constraint that the object did not extend beyond the frame.
The amount of translation per timestep varied from $-12$ to $12$ pixels for the $x$-axis translation and $y$-axis translation.
The amount of rotation per timestep varied from $4$ to $12$ degrees, where the direction was counterclockwise.

We evaluated distance between the transformed image $\mathbf{o}_{0,i}$ and the ground truth image $\mathbf{o}_i$.
We used the MSE loss in Eq.~\ref{eq:pixel_base_recons_loss} as the evaluation metric $I_r$.
The smaller $I_r$, the closer the transformed image $\mathbf{o}_{0,i}$ to the ground truth image $\mathbf{o}_i$.
The minimum value of $I_r$ was $0$.

Fig.~\ref{fig:2_a} shows the difference in the image reconstruction metric $I_r$ between the decoder without conditioning $D$ and the conditioned decoder $D_{i,j}$.
The image reconstruction metric $I_r$ with the conditioned decoder $D_{i,j}$ tended to be smaller than that with the decoder without conditioning $D$ ($p<0.001$, $t$-test).

Fig.~\ref{fig:2_b} shows an example of the transformed image $\mathbf{o}_{0, 6}$ and feature transformations $g_{0, 1}$ and $v_{0, 1}$.
The arrows represent the coordinate shifts in feature transformations $g_{0, 1}$ and $v_{0, 1}$.
Regardless of the decoder, the transformations were categorized into translations and rotations.
However, there was a difference in the quality of the transformed images $\mathbf{o}_{0, 6}$.
The decoder without conditioning $D$ failed in patch-centered rotations, which resulted in jagged contours.
By contrast, the conditioned decoder $D_{i,j}$ succeeded in patch-centered rotations, which resulted in a smooth contour.

\begin{figure}[t]\centering
\begin{subfigure}[t]{0.362\columnwidth}\centering\fbox{\includegraphics[width=0.95\textwidth]{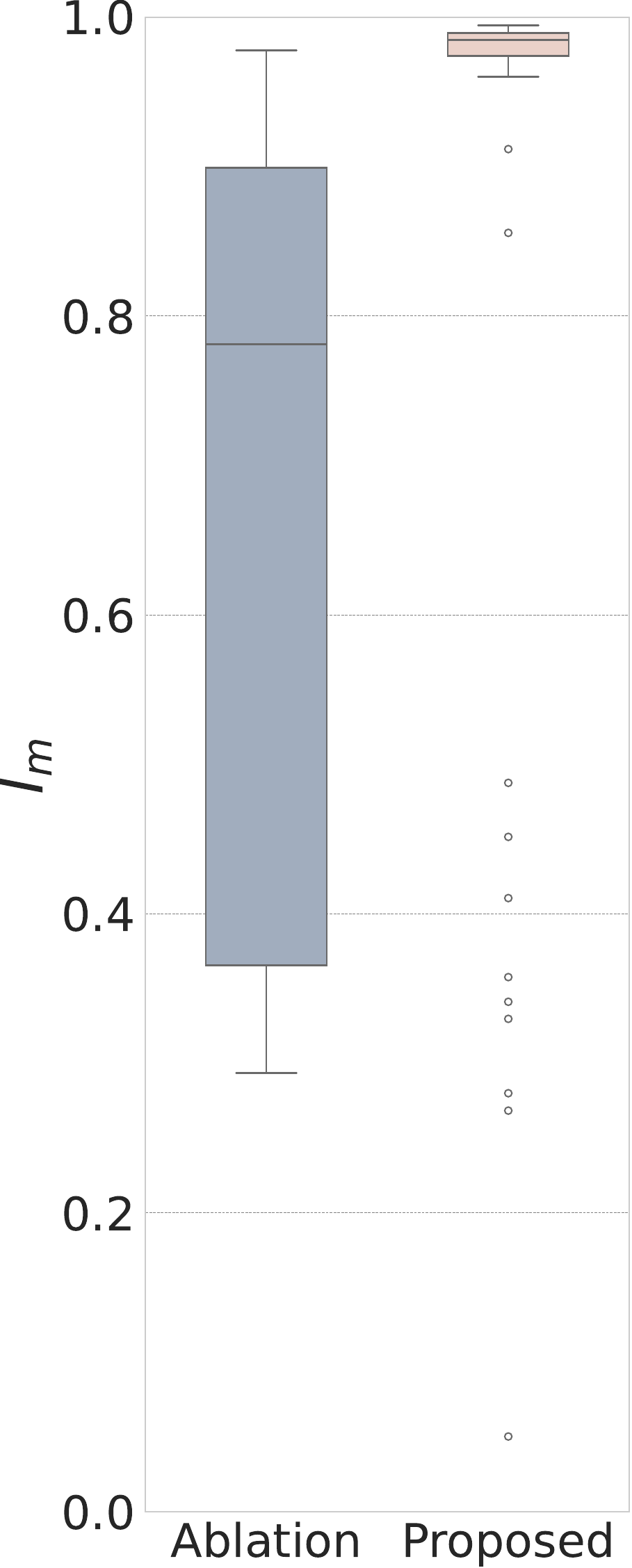}}\caption{}\label{fig:3_a}\end{subfigure}
\begin{subfigure}[t]{0.618\columnwidth}\centering\fbox{\includegraphics[width=0.95\textwidth]{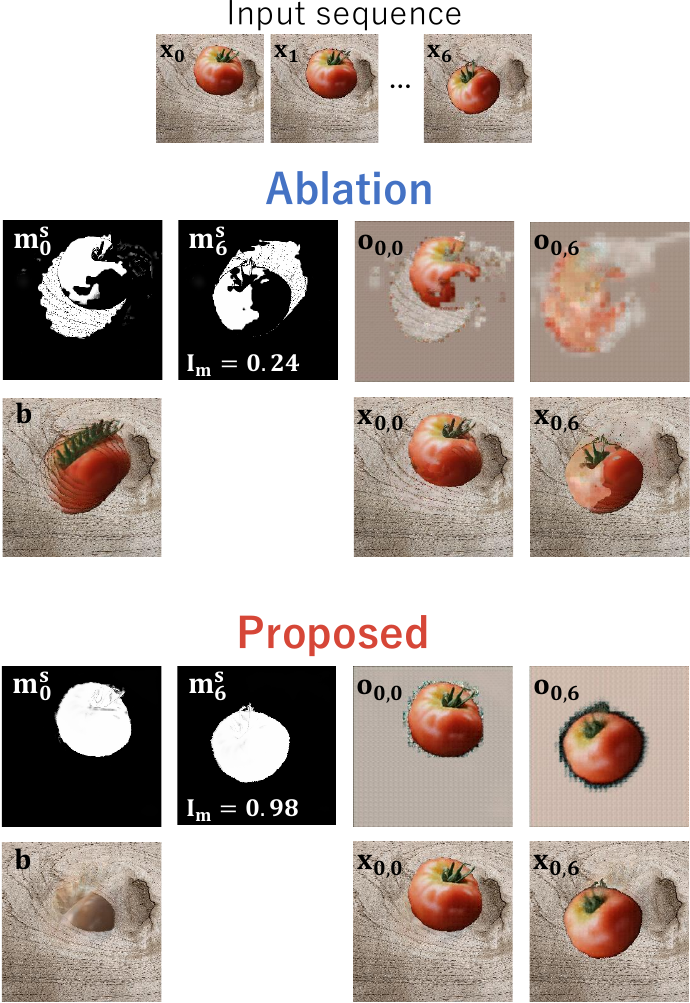}}\caption{}\label{fig:3_b}\end{subfigure}\caption{
Results of comparison experiment on \textbf{Real-obj-bg}. 
\textbf{(a)} Boxplot comparing the object segmentation metric $I_m$ with mask reconstruction loss $\mathcal{L}_m$ (proposed) and without it (ablation).
\textbf{(b)} Examples of object masks $\mathbf{m}^s_0, \mathbf{m}^s_6$, transformed object images $\mathbf{o}_{0,0}, \mathbf{o}_{0,6}$, background image $\mathbf{b}$, and transformed scene images $\mathbf{x}_{0,0}$ and $\mathbf{x}_{0,6}$.
The object segmentation metric $I_m$ is the metric in one sequence, not the average over all sequences.
}\label{fig:3}\end{figure}

\begin{figure}[t]\centering\includegraphics[width=\linewidth]{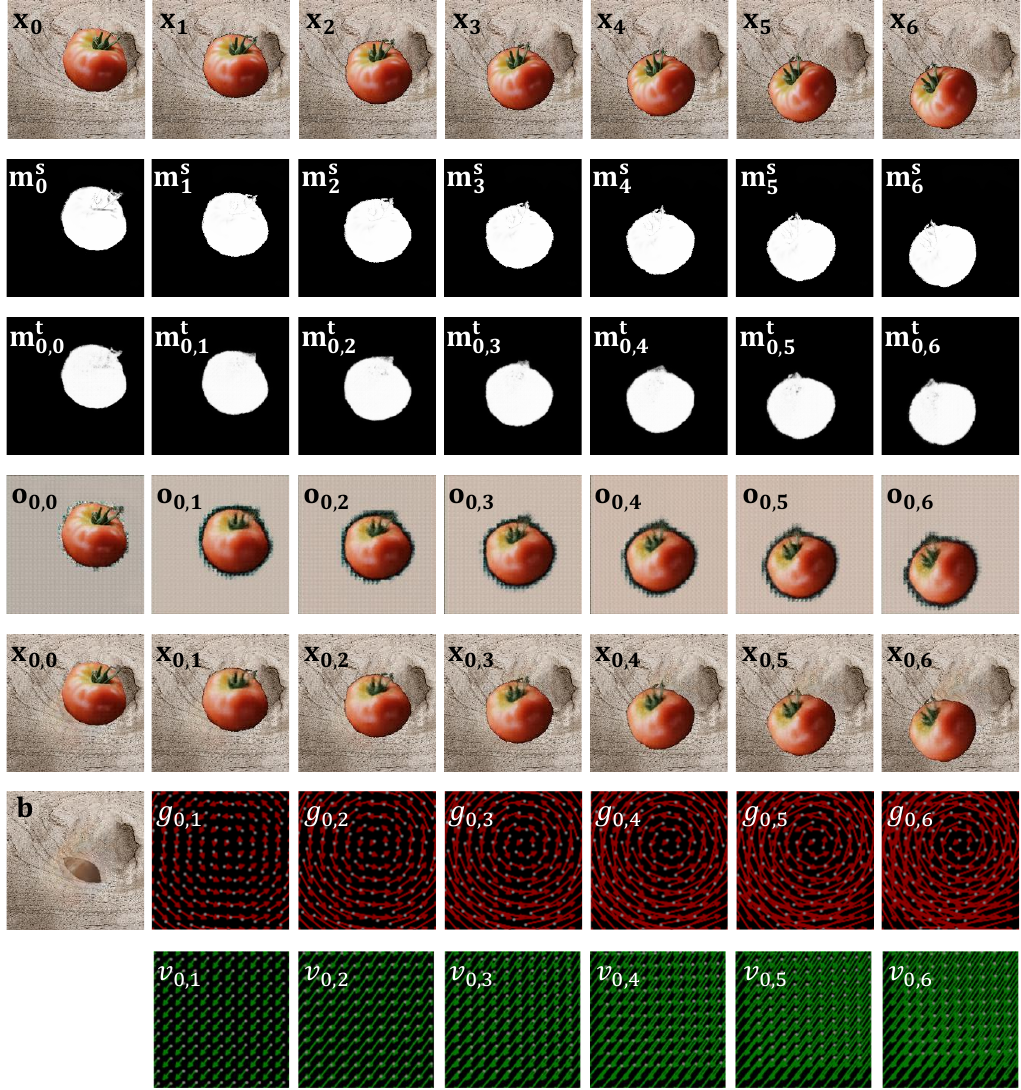}\caption{
Example of learning results in \textbf{Real-obj-bg}.
}\label{fig:4}\end{figure}

\subsection{Results for a real-world object and background}\label{sec:results_on_real_world_object_and_background}

We demonstrated the effectiveness of the mask reconstruction loss $\mathcal{L}_m$.
We compared performance when not using the mask reconstruction loss $\mathcal{L}_m$ by setting the weight $\eta$ to $0$.
As a result, we demonstrated that the proposed method simultaneously learned object segmentation and transformation categorization without supervision.
We visualized an example of learning results including the object mask $\mathbf{m}^s_i$, and categorized transformations $g_{0,i}$ and $v_{0,i}$.

The dataset \textbf{Real-obj-bg} consisted of sequences of a real-world object (tomato) and background (wood).
We created the sequence by replacing part of the background image with the object image.
The object was just large enough to fit in a $120 \times 120$ square.
Both object and background images were license-free from \url{https://www.pexels.com/}.
The other settings were the same as those in \textbf{Syn-obj}.

We evaluated distance between the object mask $\mathbf{m}^s_i$ and the ground truth mask $\mathbf{m}^{GT}_i$.
We used intersection over union (IoU) as the evaluation metric $I_m$, which is defined as follows:
\begin{equation}
    I_m = \frac{1}{T} \sum_{i=0}^{T-1} \frac{\mathbf{m}^s_i \cap \mathbf{m}^{GT}_i}{\mathbf{m}^s_i \cup \mathbf{m}^{GT}_i},
\end{equation}
where masks were pre-binarized with a threshold of $0.5$.
IoU is a popular metric for evaluating segmentation performance \cite{Ronneberger2015}.
The larger $I_m$, the closer the object mask $\mathbf{m}^s_i$ to the ground truth mask $\mathbf{m}^{GT}_i$.
The maximum value of $I_m$ was $1$.

Fig.~\ref{fig:3_a} shows the difference of the object segmentation metric $I_m$ on using mask reconstruction loss $\mathcal{L}_m$.
The object segmentation metric $I_m$ with the mask reconstruction loss $\mathcal{L}_m$ tended to be higher than that without it ($p<0.001$, $t$-test).

Fig.~\ref{fig:3_b} shows examples of the object mask $\mathbf{m}^s_i$, transformed object image $\mathbf{o}_{0,i}$, background image $\mathbf{b}$, and transformed scene image $\mathbf{x}_{0,i}$.
Without mask reconstruction loss $\mathcal{L}_m$, although the transformed scene image $\mathbf{x}_{0,i}$ matched the ground truth scene image $\mathbf{x}_i$, we obtained the inappropriate object mask $\mathbf{m}^s_i$.
The background image $\mathbf{b}$ was a merging of scene images for all time steps $\{\mathbf{x}_i\}_{i=0}^{T-1}$.
The shapes of the transformed object images $\mathbf{o}_{0,0}$ and $\mathbf{o}_{0,6}$ were different.
The shapes of the masks $\mathbf{m}^s_0$ and $\mathbf{m}^s_6$ were also different.
The object mask $\mathbf{m}^s_i$ simply selected image regions from the background image $\mathbf{b}$ and the transformed object image $\mathbf{o}_{0,i}$ based on according to whether they were useful for image reconstruction.
By contrast, with mask reconstruction loss $\mathcal{L}_m$, appropriate results were obtained successfully.

These results demonstrate that using only object masks $\{\mathbf{m}^s_i\}_{i=0}^{T-1}$ that did not involve the feature transformation $(g \circ v)$ could result in inappropriate learning, and that using masks $\{\mathbf{m}^t_{0,i}\}_{i=0}^{T-1}$ under the same feature transformation $(g \circ v)$ led to appropriate object segmentation.

Finally, we show the result of simultaneous object segmentation and transformation categorization in Fig.~\ref{fig:4}.
The object mask $\mathbf{m}^s_i$ successfully reflected the moving object.
The transformed object mask $\mathbf{m}^t_{0,i}$ appropriately matched the object mask $\mathbf{m}^s_i$.
The transformed object image $\mathbf{o}_{0,i}$ followed an appropriate shape-invariant geometric transformation.
The background image $\mathbf{b}$ appropriately reflected the stationary region of the scene.
The transformed scene image $\mathbf{x}_{0,i}$ matched the ground truth scene image $\mathbf{x}_i$.
Transformations were categorized appropriately, with transformation $g$ representing a rotation and transformation $v$ representing a translation.

These results demonstrate that the proposed method simultaneously learned object segmentation and transformation categorization on high-resolution images containing a real-world object and background without supervision.

\section{Conclusion}\label{sec:conclusion}

In this study, we proposed a novel representation learning method that combines the transformation categorization method based on group decomposition theory with feature extraction and object segmentation.
We replaced pixel-to-pixel translation with feature-to-feature translation, retaining the formulation based on group decomposition theory.
Additionally, we indicated the necessity of rotating the feature vectors, which cannot be achieved by feature translation alone; proposed a technique to rotate the feature vectors in the decoder by simply providing the decoder with transformation parameters; and experimentally demonstrated the effectiveness of this technique.
We formulated object segmentation as the grouping of features under the same feature transformation and experimentally demonstrated its effectiveness.
We verified that the proposed method simultaneously learned object segmentation and transformation categorization on high-resolution images containing a real-world object and background without supervision.
To the best of our knowledge, this is the first study that extends the Lie group transformation learning methods \cite{Takada2021, Takada2022, Takatsuki2023, Nishitsunoi2025} to feature-based transformations on data containing real-world backgrounds.

In this study, we focused on 2D scenes; we have not yet explored 3D scenes with stereo vision.
Our method extracts 2D features from 2D sensory input.
To perform transformations in 3D space, we will need to extract 3D features.
Additionally, we need to relax the formulation of the transformation from a scalar transformation amount because the degrees of freedom of rotation in 3D space is $3$.

The extension from a fixed camera to a moving camera would bring our method closer to a human development scenario in which the head and eyes move.
Our method performs object segmentation based on the assumption that the background does not change in the sensory input.
However, the appearance of the background can change when the camera itself moves.
It would be necessary to learn a representation of camera motion by extending existing formulations of object motion, and to group together, as objects, features that cannot be explained by camera motion alone.

\bibliographystyle{ieeetr}
{\small
\bibliography{bib/references}}

\begin{thebibliography}{10}

\bibitem{Takada2021}
T.~Takada, Y.~Ohmura, and Y.~Kuniyoshi, ``Unsupervised learning of
  shape-invariant lie group transformer by embedding ordinary differential
  equation,'' in {\em IEEE International Conference on Development and Learning
  (ICDL)}, pp.~1--6, 2021.

\bibitem{Nishitsunoi2024}
K.~Nishitsunoi, Y.~Ohmura, and Y.~Kuniyoshi, ``Unsupervised learning for global
  and local visual perception using navon figures,'' in {\em Cognitive Science
  Society (CogSci)}, vol.~46, 2024.

\bibitem{Bengio2013}
Y.~Bengio, A.~Courville, and P.~Vincent, ``Representation learning: A review
  and new perspectives,'' {\em IEEE transactions on pattern analysis and
  machine intelligence}, vol.~35, no.~8, pp.~1798--1828, 2013.

\bibitem{Higgins2017}
I.~Higgins, L.~Matthey, A.~Pal, C.~Burgess, X.~Glorot, M.~Botvinick,
  S.~Mohamed, and A.~Lerchner, ``Beta-vae: Learning basic visual concepts with
  a constrained variational framework,'' in {\em International Conference on
  Learning Representations (ICLR)}, 2017.

\bibitem{Chen2016}
X.~Chen, Y.~Duan, R.~Houthooft, J.~Schulman, I.~Sutskever, and P.~Abbeel,
  ``Infogan: Interpretable representation learning by information maximizing
  generative adversarial nets,'' in {\em Neural Information Processing Systems
  (NeurIPS)}, vol.~29, 2016.

\bibitem{Yang2023}
T.~Yang, Y.~Wang, Y.~Lu, and N.~Zheng, ``Disdiff: Unsupervised disentanglement
  of diffusion probabilistic models,'' in {\em Neural Information Processing
  Systems (NeurIPS)}, vol.~36, pp.~69130--69156, 2023.

\bibitem{Higgins2018}
I.~Higgins, D.~Amos, D.~Pfau, S.~Racaniere, L.~Matthey, D.~Rezende, and
  A.~Lerchner, ``Towards a definition of disentangled representations,'' {\em
  arXiv preprint arXiv:1812.02230}, 2018.

\bibitem{Ohmura2025}
Y.~Ohmura, W.~Shimaya, and Y.~Kuniyoshi, ``Unsupervised categorization of
  similarity measures,'' {\em arXiv preprint arXiv:2502.08098}, 2025.

\bibitem{Simpson2018}
A.~Simpson, ``Category-theoretic structure for independence and conditional
  independence,'' {\em Electronic Notes in Theoretical Computer Science},
  vol.~336, pp.~281--297, 2018.

\bibitem{Nishitsunoi2025}
K.~Nishitsunoi, Y.~Ohmura, T.~Komatsu, and Y.~Kuniyoshi, ``Learning
  conditionally independent transformation using normal subgroup in group
  theory,'' {\em arXiv preprint arXiv:2504.04490}, 2025.

\bibitem{Singh1999}
A.~Singh, ``The last mathematical testament of galois,'' {\em Resonance},
  pp.~93--100, 1999.

\bibitem{Takada2022}
T.~Takada, W.~Shimaya, Y.~Ohmura, and Y.~Kuniyoshi, ``Disentangling patterns
  and transformations from one sequence of images with shape-invariant lie
  group transformer,'' in {\em IEEE International Conference on Development and
  Learning (ICDL)}, pp.~54--59, 2022.

\bibitem{Takatsuki2023}
R.~Takatsuki, Y.~Ohmura, and Y.~Kuniyoshi, ``Unsupervised judgment of
  properties based on transformation recognition,'' in {\em IEEE International
  Conference on Development and Learning (ICDL)}, pp.~409--414, 2023.

\bibitem{Krizhevsky2012}
A.~Krizhevsky, I.~Sutskever, and G.~E. Hinton, ``Imagenet classification with
  deep convolutional neural networks,'' in {\em Neural Information Processing
  Systems (NeurIPS)}, vol.~25, 2012.

\bibitem{Hendrycks2019}
D.~Hendrycks and T.~Dietterich, ``Benchmarking neural network robustness to
  common corruptions and perturbations,'' in {\em International Conference on
  Learning Representations (ICLR)}, 2019.

\bibitem{Chen2018}
R.~T. Chen, Y.~Rubanova, J.~Bettencourt, and D.~K. Duvenaud, ``Neural ordinary
  differential equations,'' in {\em Neural Information Processing Systems
  (NeurIPS)}, vol.~31, 2018.

\bibitem{Hochreiter1997}
S.~Hochreiter and J.~Schmidhuber, ``Long short-term memory,'' {\em Neural
  computation}, vol.~9, no.~8, pp.~1735--1780, 1997.

\bibitem{Ronneberger2015}
O.~Ronneberger, P.~Fischer, and T.~Brox, ``U-net: Convolutional networks for
  biomedical image segmentation,'' in {\em Medical Image Computing and
  Computer-Assisted Intervention (MICCAI)}, pp.~234--241, 2015.

\bibitem{Ulyanov2016}
D.~Ulyanov, A.~Vedaldi, and V.~Lempitsky, ``Instance normalization: The missing
  ingredient for fast stylization,'' {\em arXiv preprint arXiv:1607.08022},
  2016.

\bibitem{Burgess2019}
C.~P. Burgess, L.~Matthey, N.~Watters, R.~Kabra, I.~Higgins, M.~Botvinick, and
  A.~Lerchner, ``Monet: Unsupervised scene decomposition and representation,''
  {\em arXiv preprint arXiv:1901.11390}, 2019.

\bibitem{Komatsu2024}
T.~Komatsu, Y.~Ohmura, and Y.~Kuniyoshi, ``Ablation study to clarify the
  mechanism of object segmentation in multi-object representation learning,''
  in {\em IEEE International Conference on Development and Learning (ICDL)},
  pp.~1--7, 2024.

\end{thebibliography}

\end{document}